\PassOptionsToPackage{unicode}{hyperref}
\PassOptionsToPackage{hyphens}{url}
\PassOptionsToPackage{dvipsnames,svgnames*,x11names*}{xcolor}
\documentclass[
  11pt,
]{article}
\usepackage{lmodern}
\usepackage{amsmath}
\usepackage{ifxetex,ifluatex}
\ifnum 0\ifxetex 1\fi\ifluatex 1\fi=0 
  \usepackage[T1]{fontenc}
  \usepackage[utf8]{inputenc}
  \usepackage{textcomp} 
  \usepackage{amssymb}
\else 
  \usepackage{unicode-math}
  \defaultfontfeatures{Scale=MatchLowercase}
  \defaultfontfeatures[\rmfamily]{Ligatures=TeX,Scale=1}
\fi
\IfFileExists{upquote.sty}{\usepackage{upquote}}{}
\IfFileExists{microtype.sty}{
  \usepackage[]{microtype}
  \UseMicrotypeSet[protrusion]{basicmath} 
}{}
\makeatletter
\@ifundefined{KOMAClassName}{
  \IfFileExists{parskip.sty}{%
    \usepackage{parskip}
  }{
    \setlength{\parindent}{0pt}
    \setlength{\parskip}{6pt plus 2pt minus 1pt}}
}{
  \KOMAoptions{parskip=half}}
\makeatother
\usepackage{xcolor}
\IfFileExists{xurl.sty}{\usepackage{xurl}}{} 
\IfFileExists{bookmark.sty}{\usepackage{bookmark}}{\usepackage{hyperref}}
\hypersetup{
  pdftitle={Panarchy: ripples of a boundary concept},
  colorlinks=true,
  linkcolor=blue,
  filecolor=Maroon,
  citecolor=blue,
  urlcolor=blue,
  pdfcreator={LaTeX via pandoc}}
\usepackage{graphicx}
\makeatletter
\def\maxwidth{\ifdim\Gin@nat@width>\linewidth\linewidth\else\Gin@nat@width\fi}
\def\maxheight{\ifdim\Gin@nat@height>\textheight\textheight\else\Gin@nat@height\fi}
\makeatother
\setkeys{Gin}{width=\maxwidth,height=\maxheight,keepaspectratio}
\makeatletter
\def\fps@figure{htbp}
\makeatother
\usepackage{dcolumn, rotating, longtable, lineno, float, array, tabularx}

\usepackage[margin=2.5cm]{geometry}
\ifluatex
  \usepackage{selnolig}  
\fi
\newlength{\cslhangindent}
\newlength{\csllabelwidth}
\newenvironment{CSLReferences}[2] 
 {
  \setlength{\parindent}{0pt}
  \ifodd #1 \everypar{\setlength{\hangindent}{\cslhangindent}}\ignorespaces\fi
  \ifnum #2 > 0
  \setlength{\parskip}{#2\baselineskip}
  \fi
 }%
 {}
\usepackage{calc}

\title{Panarchy: ripples of a boundary concept}
\author{\small Juan Rocha\textsuperscript{1,2,3}, Linda
Luvuno\textsuperscript{4}, Jesse Rieb\textsuperscript{5}, Erin
Crockett\textsuperscript{6}, Katja Malmborg\textsuperscript{1}, Michael
Schoon\textsuperscript{7}, Garry Peterson\textsuperscript{1}\\
\small\\
\footnotesize \textsuperscript{1}Stockholm Resilience Centre, Stockholm
University, Kräftriket 2B, 10691 Stockholm\\
\footnotesize \textsuperscript{2}Future Earth, The Swedish Royal Academy
of Sciences, Lilla Frescativägen 4A, 104 05 Stockholm\\
\footnotesize \textsuperscript{3}South American Institute for Resilience
and Sustainability Studies, Maldonado, Uruguay\\
\footnotesize \textsuperscript{4}Center for Complex Systems in
Transition, Stellenbosch University, South Africa\\
\footnotesize \textsuperscript{5}Department of Geography, McGill
University, 805 rue Sherbrooke ouest, Montréal (Québec) H3A 0B9,
Canada\\
\footnotesize \textsuperscript{6}Department of Natural Resource
Sciences, McGill University, 21111 Lakeshore Road,\\
\footnotesize Sainte-11 Anne-de-Bellevue, QC, H9X 3V9, Canada\\
\footnotesize \textsuperscript{7}School of Sustainability, Arizona State
University, Tempe, Arizona, USA}
\date{}

\begin{document}
\maketitle
\begin{abstract}
How do social-ecological systems change over time? In 2002 Holling and
colleagues proposed the concept of Panarchy, which presented
social-ecological systems as an interacting set of adaptive cycles, each
of which is produced by the dynamic tensions between novelty and
efficiency at multiple scales. Initially introduced as a conceptual
framework and set of metaphors, panarchy has gained the attention of
scholars across many disciplines and its ideas continue to inspire
further conceptual developments. Almost twenty years after this concept
was introduced we review how it has been used, tested, extended and
revised. We do this by combining qualitative methods and machine
learning. Document analysis was used to code panarchy features that are
commonly used in the scientific literature (N = 42), a qualitative
analysis that was complemented with topic modeling of 2177 documents. We
find that the adaptive cycle is the feature of panarchy that has
attracted the most attention. Challenges remain in empirically grounding
the metaphor, but recent theoretical and empirical work offer some
avenues for future research.
\end{abstract}

\hypertarget{introduction}{%
\section{Introduction}\label{introduction}}

Almost two decades ago, the edited book ``Panarchy: Understanding
Transformations in Human and Natural Systems'' (Gunderson and Holling
2002), presented a synthetic perspective of how a group of leading
social-ecological researchers associated with the Resilience Alliance
understood change in social-ecological systems. The concept of Panarchy
was a key focus on this influential book. In the late 1990s and early
2000s, the Resilience Alliance was a productive, innovative highly
collaborative group of environmental scientists that was focussed on
bridging social and natural sciences as well as theory and practice.
They were very focussed on addressing social-ecological problems by
combining insights from the social and natural sciences, as well as the
arts and humanities (Parker and Hackett 2012).

Panarchy encapsulates a set of concepts that have inspired the work of
environmental scientists and practitioners for almost two decades. It
was a central concept produced by the Resilience Alliance, an
international group of influential academics focused on the resilience
of social-ecological systems in theory and practice (Parker and Hackett
2012). The Panarchy concept builds on Holling's adaptive cycle (C.
Holling 1986) by extending the idea across spatial and temporal scales.

Panarchy remains a boundary object that has inspired research topics,
enabled collaborations and nurtured new scientific frameworks. The ideas
put forward have been applied in field studies, archaeology,
mathematical models, participatory work, and scenario development.
Panarchy has inspired resilience assessments and guided decision making.
In this article we pay tribute to the book by studying how the concepts
and metaphors proposed have been further developed in the academic
literature. We also document criticism of the concepts and identify key
research frontiers.

Panarchy proposes that it is useful to conceptualize systems in terms of
interacting adaptive cycles. The adaptive cycle was an idea first
proposed by Holling based upon his experience working and studying
managed ecosystems (C. Holling 1986). It was meant to be a conceptual
tool that focussed attention on processes of destruction and
reorganization, which had been neglected in comparison to those of
growth and conservation. Considering these processes provides a more
complex may to understand system dynamics.

An adaptive cycle alternates between long periods of system aggregation,
connection, and accumulation, and shorter periods of disruption and
reorganization. The adaptive cycle exhibits two major phases. The first,
often referred to as the front-loop, from r to K, is the slow,
incremental phase of growth and accumulation. The second, referred to as
the back-loop, from Omega to Alpha, is the rapid phase of reorganization
leading to renewal. A system going into the back-loop can either remerge
in a similar form, or may re-emerge as a new type of system with new
boundaries and key components. Resilience researchers proposed that
tensions between demands for organization and efficiency versus demands
for novelty and diversity drive adaptive cycle type dynamics in many
different types of complex systems (C. S. Holling, Gunderson, and
Peterson 2002).

Panarchy is not a theory of what it is, but a metaphor of what might be
(Gunderson and Holling 2002). It is not a predictive tool, but aims to
understand adaptive change. The adaptive cycle is hypothesized to exist
within a three-dimensional space defined by three properties: potential,
connectivity, and resilience (Gunderson and Holling 2002). Potential
refers to the capital available capital to the system, for example
nutrients and carbon captured by a forest, or human capital - skills and
knowledge accumulated to run the economy. Connectivity is a proxy of the
structure of the system; it is the network of interactions and strength
between its elements. Resilience is the capacity of any system to absorb
disturbance and reorganize while undergoing change so as to retain
essentially the same identity, as determined by its function, structure,
and feedbacks (Folke 2016).

\begin{figure*}[ht]
\centering
\includegraphics[width = 6in, height = 2.5in]{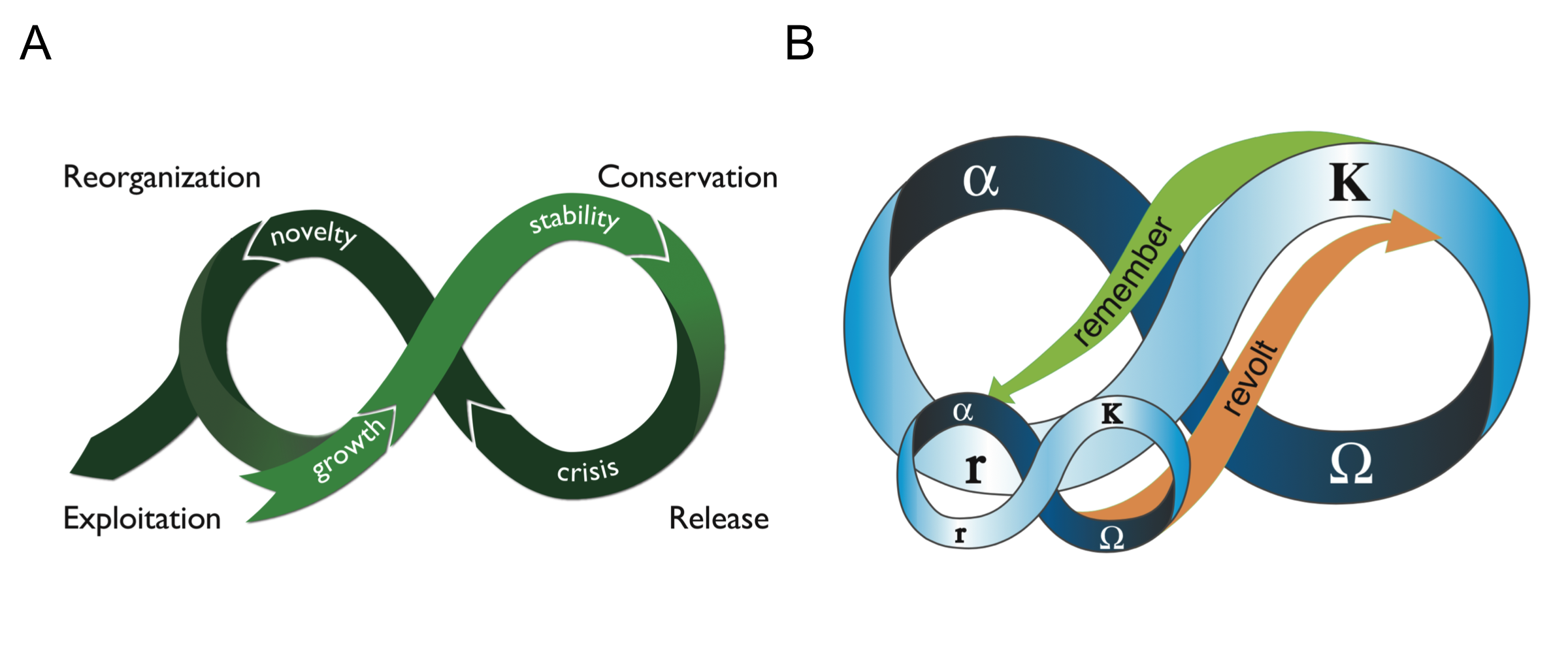}
\caption{\textbf{Panarchy} is an heuristic of nested adaptive cycles that serves to represent a variety of systems and environmental problems. Adaptive cycles (A) at different scales of the hierarchy (B) can be connected through remember and revolt cross-scale interactions.}
\label{fig:fig1}
\end{figure*}

Panarchy posits that systems are organized in nested hierarchies across
space and time, where each level of the hierarchy is a subsystem that
can be in a different phase of the adaptive cycle. These phases are
exploitation, conservation, release, and reorganization, the latter
characterized by events of creative destruction (Fig. \ref{fig:fig1}).
They are inspired by mathematical models used in economics and ecology,
but broadly describe patterns of growth, collapse, and recovery that are
common to populations, ecological communities, markets, or political
organizations. While each subsystem in the hierarchy can be at a
different phase of the cycle, these subsystems can influence each other
through cross-scale interactions called ``revolt'' and ``remember.''

The three-dimensional space has corners with attractors that can cause
the adaptive cycle to stagnate: poverty and rigidity traps (Gunderson
and Holling 2002). Poverty traps are described in panarchy as
maladaptive states where potential, connectivity and resilience are low.
Poverty traps are a series of feedback mechanisms that reinforce
impoverished states (Bowles, Durlauf, and Hoff 2006), limiting the
system's capacity to innovate and increase potential. The opposite
corner, where potential, connectivity and resilience are high, is
another maladaptive space called a rigidity trap. In that corner there
is little space for experimentation and innovation. Examples include
systems where ecological resilience has been expensively replaced by
artificial processes to maintain the system such as levees, flood
barriers, or chemical control of pests (C. S. Holling and Meffe 1996).

Panarchy offers a rich conceptual framework for understanding
environmental problems. While inspired by a few mathematical constructs
(e.g.~cycles, traps, scaling laws), it is general enough to invite
scholars from multiple disciplinary backgrounds, ontologies and
epistemologies, to collaborate around research questions and applied
problems. As such, it is useful as a boundary object and can be used
empirically or metaphorically. The book itself presents a series of case
studies where geographers, economists, political scientists, and
ecologists have demonstrated the utility of the framework to their area
of research. However, the extent to which these concepts have gained
empirical support beyond the metaphor, or how they have evolved over
time, remain open questions. For example, the adaptive cycle is a useful
metaphor to look back at history and organize events and periods around
phases, but it remains challenging to identify the phase of a system in
real time, or to use panarchy to project future trajectories. In other
words, panarchy is useful for retrospective studies, but its
applications may be more limited for prospective ones. Are we at a stage
where we can start distilling theory that enables case study
comparisons, deriving functional forms, or draw theoretical expectations
and predictions?

Here we review the academic literature of the last two decades to trace
how these concepts have evolved, in what type of problems they have been
applied and found useful, and finally what remains as the key frontiers
of research.

\hypertarget{methods}{%
\section{Methods}\label{methods}}

To answer these questions we combined an automated literature review
based on topic modeling (Blei 2012; Griffiths and Steyvers 2004) with
human-coded document analysis (Bryman 2008). All data and code to
replicate the analysis are available at:
\url{https://github.com/juanrocha/panarchy} and a public online
repository (\url{10.6084/m9.figshare.13490919}).

\emph{Data}: We used the Web of Science, Scopus, and GoogleScholar to
survey academic literature that has used or referenced works that trace
back to Panarchy (Gunderson and Holling 2002). We extracted complete
records from the Scopus database that matched the search for ``(panarchy
OR adaptive cycle)'' (N = 595), or the search ``(panarchy OR adaptive
cycle) AND resilience'' (N = 278). The data was combined with all papers
(N = 1923) that cited the inaugural paper that introduced the book to
the academic community (C. S. Holling 2001). Records with missing
abstracts were dropped (N = 191), and records with missing year were set
to 2020 given that they are accepted manuscripts with digital object
identifier (doi) scheduled to be published later in 2020 or 2021.

To prepare the data for topic modeling, we constructed a document term
matrix with documents in rows and words in columns. Here our unit of
analysis for the document are the abstracts retrieved, and the matrix
contains the count of words per document. We removed stop words
(e.g.~``the,'' ``a'') and digits from the matrix, as well as a list of
words that were over represented in our data and are common in the
scientific literature but are unrelated to the papers' topics
(e.g.~``paper,'' ``study,'' ``aim'').

\emph{Topic models}: are an unsupervised statistical technique to reduce
the dimensionality of a corpus of data (typically but not necessarily
text) into topics (Blei 2012). Here a topic is a latent variable that
ranks words with high probability of appearing together within the same
document. Documents in turn can be described by the probability
distribution of a particular set of topics. Since they are (posterior)
probabilities, the sum of the probability of all words for any given
topic should be one, and the sum of the probability of all topics for
any given document should also be one. An iterative process or algorithm
is what allows the model to learn the ranking of words that best explain
topics, and the ranking of topics that best explain documents.

The underlying statistical technique for this machine learning approach
is called Latent Dirichlet Allocation (LDA) (Blei, Ng, and Jordan 2003).
It allocates probabilities to latent variables (topics) based on the
distribution of words in text data, assuming a multivariate continuous
(Dirichlet) distribution. We compared three LDA algorithms: correlated
topic models (CTM), variational expectation maximization (VEM), and
Gibbs sampling (Gibbs), by assessing their performance against their
log-likelihood estimation, entropy and perplexity (Grün and Hornik
2011). Entropy is a measure of order or disorder of a system; in the
context of topic models it measures how evenly the topic distribution is
spread. Perplexity measures the uncertainty of predicting a single word,
so if the model performance were the same as random, perplexity would
approximate the vocabulary size (16 728 words). These metrics enable us
to choose which algorithm best fits our data, what is the optimal number
of topics to fit, and how to avoid overfitting.

\emph{Document analysis}: We complemented our unsupervised approach with
document analysis (Bryman 2008) by coding an additional set of
categorical variables for a sample of highly cited papers (N = 41). We
annotated qualitative aspects such as use of the adaptive cycle,
identification of its phases, whether the paper is conceptual, modeling
or an empirical study. When empirical, we record the temporal and
spatial scales of the case study. We also identified what aspects of
panarchy were most used in the papers. For example whether there is
emphasis on cross-scale interactions, or poverty and rigidity traps. We
used text annotations to capture potential criticisms as well as the
methods used. The code book is available in the supplementary material
(archived in \url{10.6084/m9.figshare.13490919}).

\hypertarget{results}{%
\section{Results}\label{results}}

Panarchy, the book published in 2002, has been cited over 7200 times
according to Google Scholar. The scientific paper that introduced the
book to the scientific community (C. S. Holling 2001) has received 1715
citations in the Web of Science, and 1923 in Scopus. Roughly half of the
citations have come from environmental (28\%) and social sciences
(22\%). Computer science (2.1\%) and arts and humanities (2.6\%) have
been the least represented.

\begin{figure*}[ht]
\centering
\includegraphics[width = 5in, height = 4in]{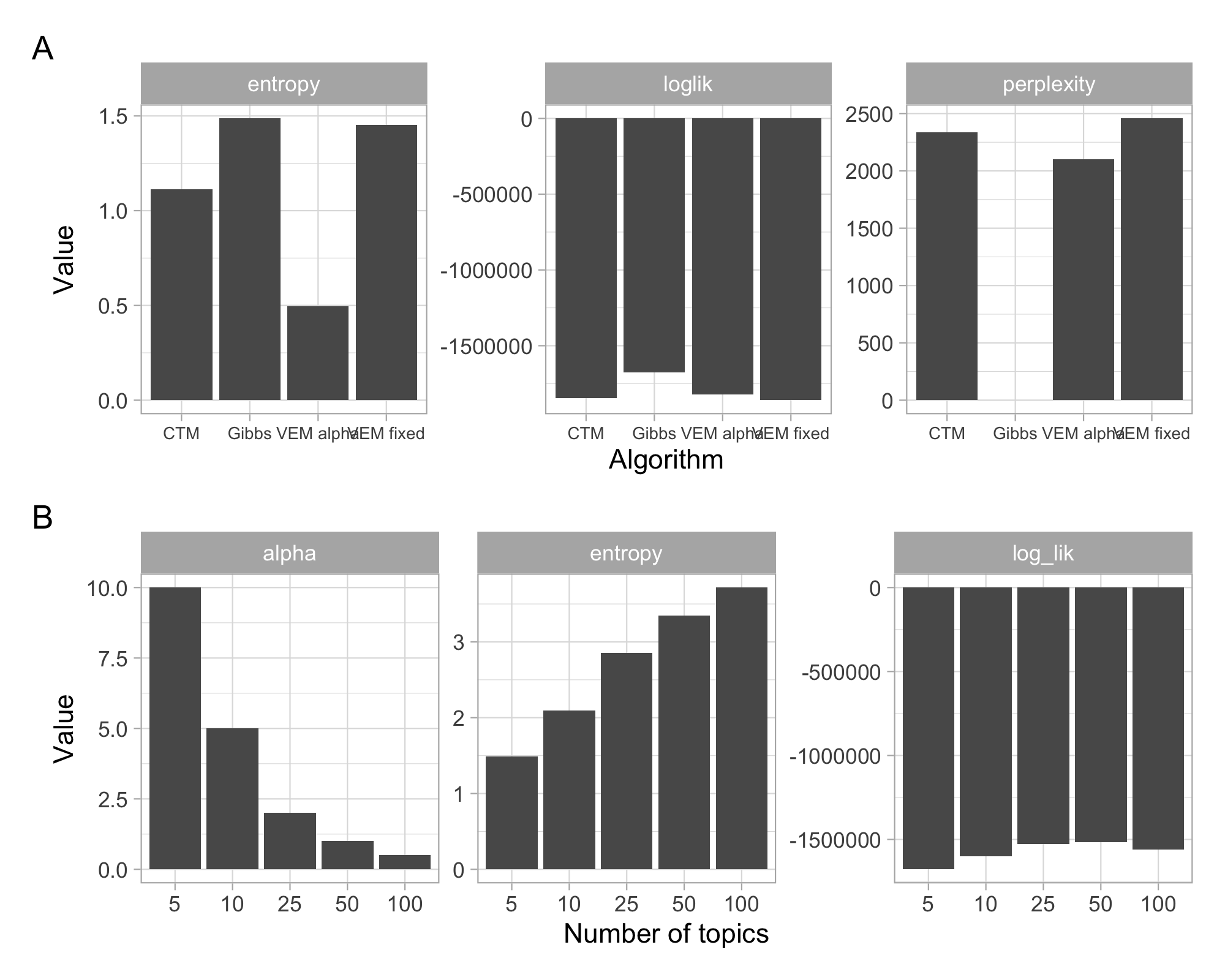}
\caption{\textbf{Algorithm and number of topic selection} Gibbs sampling maximizes entropy and log-likelihood estimation, making it a suitable algorithm for our data (A). Increasing the number of topics (from 5-100) shows that $\alpha$ decreases, suggesting that despite the larger number of topics, a few of them suffice to describe most papers (B). Log-likelihood is maximized for 50 topics followed closely with 25. While 50 topics is marginally better, for visualization purposes we restrict our analysis to 25. Note that perplexity cannot be calculated for Gibbs sampling, hence the missing value in A and absence in B.}
\label{fig:algorithm}
\end{figure*}

Comparing algorithm performance revealed that Gibbs sampling was the
best fit for the data (Fig \ref{fig:algorithm}). As a rule of thumb, an
ideal method should maximize entropy and the log-likelihood estimation
while minimizing perplexity (Grün and Hornik 2011). Gibbs sampling
maximised both entropy and likelihood with our data when compared to
other alternatives. The second best was the variational-expectation
maximisation algorithm (VEM) when \(\alpha\) was not set constant.
\(\alpha\) is a hyper parameter that weights the evenness of topic
distribution. A lower \(\alpha\) than default values indicates that the
documents can be described by rather fewer topics, or that its
distribution is very uneven. In fact, we observed that increasing the
number of topics from 5 to 100 topics did increase entropy at expense of
reducing \(\alpha\), meaning that despite the larger number of topics,
the main content of a document is still captured by a few of them. The
log-likelihood maximization stopped at around 25-50 topics. Thus, we
restricted the rest of our analysis to 25 topics.

\begin{figure*}[ht]
\centering
\includegraphics[width = 5in, height = 5in]{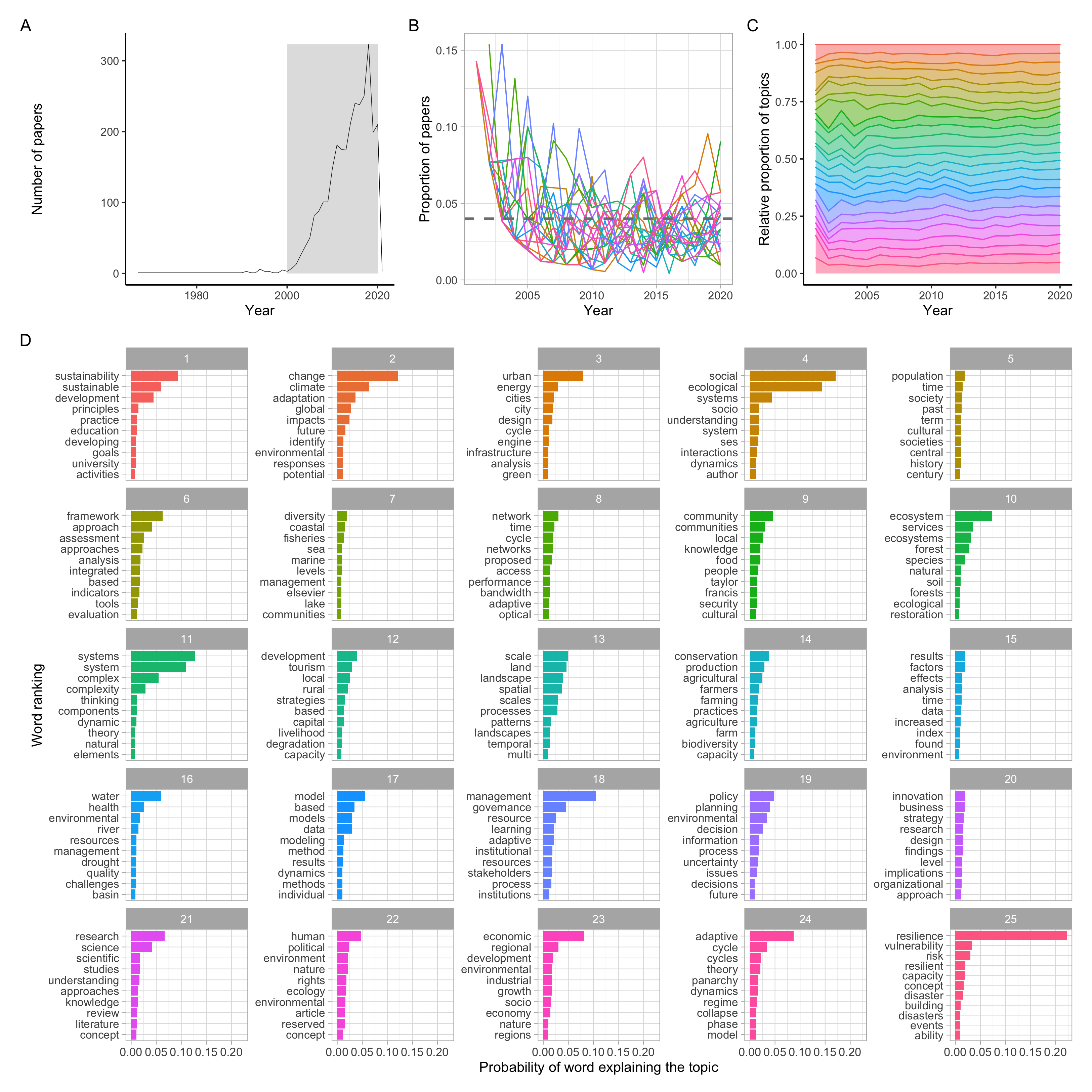}
\caption{\textbf{Panarchy topics over time} The number of papers per year (A) with a maximum of 324 in 2019. The proportion of papers per year (B) and the relative proportion of topic content per year (C) do not show strong trends for the time window with most papers (gray area in A). Each topic is summarized in (D) by the top 10-words that best describe them according to the posterior probability of our model fit. See FigS1 for a panel of time series per topics}
\label{fig:topics25}
\end{figure*}

A topic is a set of words that are ranked according to a probability
that they represent an underlying content of the document (Fig
\ref{fig:topics25}). For example, the words ``resilience,''
``vulnerability,'' ``risk'' and ``disaster'' have a high probability to
capture the content of topic 25. Papers early in our time series
(2001-3) have high content largely dominated by topics 25 on resilience,
24 on the adaptive cycle, and 18 on adaptive governance and management.
Towards the end of the time series, topic 3 on urban systems and 9 on
local communities and knowledge have spikes up to 9\% of the content of
each year (2019-2020). For comparison, if all topics were equally
represented in the content, they would have 4\% in the corpus (grey line
in Fig \ref{fig:topics25}B).

Despite fluctuations, most topics showed a relatively constant level of
interest over time (Fig \ref{fig:topics25}, Fig \ref{fig:figS1}). We did
not observe strong trends, but some topics have gained a small amount of
attention. For example, topic 1 on sustainability, or topic 10 on
ecosystem services appeared consistently across time. In contrast,
research on business innovation (topic 20), urban infrastructure (topic
3), or archaeology (topic 5) have gained attention in recent years.
Topic 8 was an outlier, with a selection of papers ranking high on
content related to network infrastructure and performance, possibly from
engineering disciplines. It was the only topic with a set of papers that
was clearly distinct from the rest of the collection, and showed a
decreasing trend over time.

\begin{figure*}[ht]
\centering
\includegraphics[width = 5in, height = 2.5in]{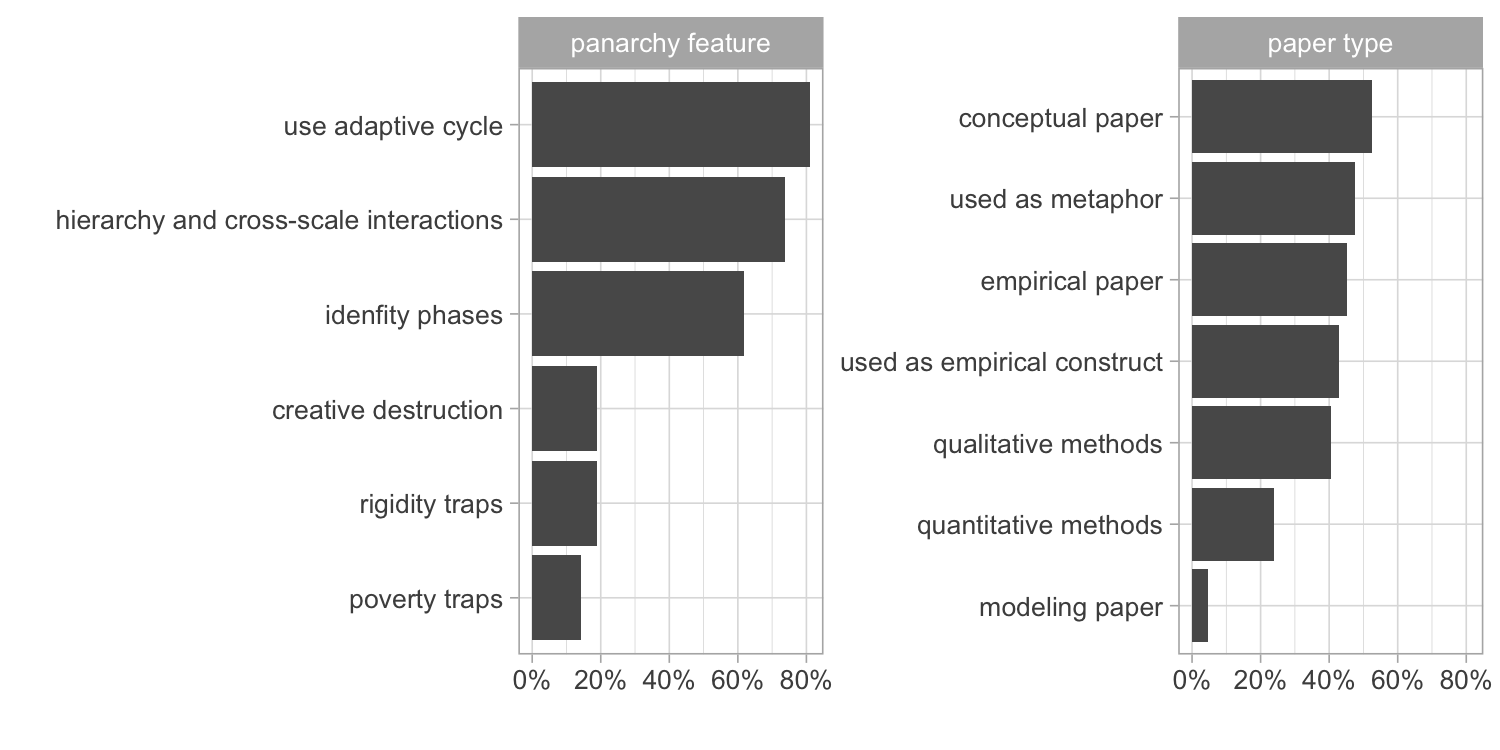}
\caption{\textbf{Qualitative results} Document analysis was used to disentangle the different panarchy features addressed by a sub-sample of papers (N = 42). Over half of the papers are conceptual work, and most empirical papers fall into qualitative methods.}
\label{fig:qual}
\end{figure*}

The human-coded document analysis revealed that the most common feature
of panarchy in the literature was the adaptive cycle (81\%, N = 42)
followed by cross-scale interactions (Fig \ref{fig:qual}). Poverty and
rigidity traps were less studied features in our sample, even when
accounting for slightly different terminology such as ``lock-ins.'' The
bulk of the papers analyzed were conceptual papers, many without a
method or discernible research question. Roughly half of the papers were
conceptual or only used panarchy as a metaphor. Over 40\% of papers in
our sample went a step further and used panarchy as an empirical
construct, for example by attempting to identify the phases of the
adaptive cycle (61\%). Out of the 22 empirical cases, 6 were at the time
scale of centuries, 7 on decades, 6 on years, and 1 in weeks. Spatially,
4 were at a city scale, 17 regional, and 1 national. Empirical papers
were dominated by qualitative methods (77\%, N = 22) and were generally
retrospective historical reviews.

The high level of conceptualization but lack of empirically grounded
hypotheses or theory testing came across as one of the major
limitations. Nonetheless, several of the papers reviewed attempted to
identify the adaptive cycle either through qualitative or quantitative
methods. Identification of adaptive cycles has found applications in a
wide range of disciplines and research problems, from delimitation of
periods in archaeology and anthropology (Redman and Kinzig 2003), to
financial crises in Europe (Castell and Schrenk 2020), or traffic jams
in China (Zeng et al. 2020). Recent empirical tests of the adaptive
cycle innovate on the types of data and methods that one can use.
Information transfer methods based on entropy have been suggested to
approximate relevant components of a system and the empirical proxies of
the axis where the adaptive cycle unfolds: potential, connectedness, and
resilience (Castell and Schrenk 2020). Percolation methods combined with
big data have been shown to be useful to test hypotheses of regime
shifts in urban systems (originally proposed in Panarchy (Gunderson and
Holling 2002)), and derive the temporal and spatial scales at which the
adaptive cycle emerges (Zeng et al. 2020).

\hypertarget{discussion}{%
\section{Discussion}\label{discussion}}

Given the strong dominance of conceptual papers, it is not surprising
that one of the key frontiers of research is to empirically ground such
conceptualizations in a way that one can test hypotheses and advance
theory beyond metaphors. Metaphors have played an important role as
boundary objects enabling interdisciplinary dialogues. We observe it on
the wide applications that panarchy has inspired, from anthropology to
engineering. In the empirical realm, the adaptive cycle is the panarchy
dimension that researchers engage the most with, but it is dominated by
qualitative and retrospective studies. That means, we still lack the
theory and methods to gather observations and be able to decide in which
phase of the adaptive cycle a system currently is, or what is the
probability of it transitioning to a new phase?

The criticism of over conceptualization is not unique to the literature
engaging Panarchy ideas. A recent review of sustainability science
mapped the different schools of thought that the discipline has
developed over the last decades (Clark and Harley 2020). A similar
conclusion was reached, where too many conceptual frameworks have been
developed, but fewer empirical attempts try to test the frameworks
against data to falsify hypotheses. The review emphasizes the problem of
measurement and observation (Clark and Harley 2020). In the context of
panarchy, recent work has developed an information theory based approach
that enables the identification of adaptive cycles (Castell and Schrenk
2020). The authors identify phases of the adaptive cycle in the European
financial crises and in grassland ecosystems, but fail to find support
for the difference in speed between the forward and backward loops
originally proposed in Panarchy.

The panarchy dimensions that received less attention in our qualitative
analysis are creative destruction, rigidity and poverty traps. This may
be at least partly due to the scope of our data: papers that have cited
Panarchy or Holling's (Gunderson and Holling 2002; C. S. Holling 2001).
Concepts such as creative destruction and poverty traps precede Panarchy
and therefore have been theorized and empirically grounded outside the
panarchy stream of thinking. For example, the theory of poverty traps
dates back to the 1950's in economics, and has received both theoretical
development (Bowles, Durlauf, and Hoff 2006) as well as empirical
grounding (Banerjee et al. 2015; Banerjee and Duflo 2012) that has
enabled researchers and governments to distinguish what kind of
interventions are likely to reduce poverty. Here we encourage similar
empirical efforts in formalizing theory and empirical support for the
adaptive cycles, their cross-scale interactions (remember and revolt),
or how the hierarchical nested nature of complex systems enhance or
erode resilience in social-ecological systems.

This paper aimed to study the use of panarchy related concepts and their
evolution since the publication of the book. Systematic literature
reviews often suffer from the limitation of restricting sample size to a
subset of what is readable on the time frame of a project -- between
dozens and maybe hundreds of papers. Qualitative analysis offers rich
insights into the papers but is limited by sample size. In our analysis,
the qualitative insights could be biased towards a non-random selection
of papers that aligned our research interests, or highly cited, review
type of work. Topic modeling enabled us to complement the analysis to
all papers reported in major scientific databases. It has the advantage
of reproducibility and reduces sample bias, but offers little insights
about the dimensions of panarchy used, methods, or criticisms. Here we
show that both methods combined are a powerful combination. Future
studies could benefit from including gray literature such as theses,
books, non-governmental organizations reports, government agencies
reports, or non-english literature. Previous studies have shown that the
use of full text instead of abstracts can also offer additional insights
on the automated analyses, for example in attributing impacts of
ecosystem services from regime shifts in social-ecological systems
(Rocha and Wikström 2015). Replication attempts do however face the
challenge of accessing full text when many of the papers are behind
pay-walls.

\hypertarget{conclusion}{%
\section{Conclusion}\label{conclusion}}

We reviewed the academic literature of the last two decades to trace how
Panarchy related concepts have evolved, in what type of problems they
have been applied and found useful, and what remains as the key
frontiers of research. Despite a growing body of literature, no topic
seems to dominate the academic production of scholars using these
concepts. The feature most used is the adaptive cycle, and the problems
where it is found most useful is in studies with a historical
perspective, either in short time horizons such as natural resource
management, urban development, or conservation; all the way to
archeology and anthropology studies on the time horizon of centuries to
millennia. Hierarchies in scale or maladaptive traps have received
comparatively less attention. The frontiers of research point out to the
problem of measurement and prediction. For example, how to observe and
approximate resilience, connectedness and potential in the present; or
how likely a current social-ecological system is to undergo a transition
in the phases of the adaptive cycle, or a ``memory'' or ``revolt'' type
of cross-scale interaction. While our survey of the literature offers a
few options to empirically ground panarchy concepts, we believe the
problem of measurement and prediction are fertile ground for future
research efforts.

\hypertarget{acknowledgements}{%
\section{Acknowledgements}\label{acknowledgements}}

JCR acknowledges financial support from Formas research grant
(942-2015-731), the Stockholm Resilience Centre, and Stockholm
University.

\hypertarget{references}{%
\section{References}\label{references}}

\hypertarget{refs}{}
\begin{CSLReferences}{1}{0}
\leavevmode\hypertarget{ref-Banerjee:2012tq}{}%
Banerjee, Abhijit, and Esther Duflo. 2012. \emph{{Poor Economics}}. A
Radical Rethinking of the Way to Fight Global Poverty. Hachette UK.

\leavevmode\hypertarget{ref-Banerjee:2015ja}{}%
Banerjee, Abhijit, Esther Duflo, Nathanael Goldberg, Dean Karlan, Robert
Osei, William Parienté, Jeremy Shapiro, Bram Thuysbaert, and Christopher
Udry. 2015. {``{A multifaceted program causes lasting progress for the
very poor: Evidence from six countries}.''} \emph{Science} 348 (6236):
1260799--99.

\leavevmode\hypertarget{ref-Blei:2012dk}{}%
Blei, David M. 2012. {``{Probabilistic topic models}.''} \emph{Commun
ACM} 55 (4): 77--84.

\leavevmode\hypertarget{ref-Blei:2003tn}{}%
Blei, David M, Andrew Y Ng, and Michael I Jordan. 2003. {``{Latent
dirichlet allocation}.''} \emph{J. Mach. Learn. Res.} 3: 993--1022.

\leavevmode\hypertarget{ref-Bowles:2006hx}{}%
Bowles, Samuel, Steven N Durlauf, and Karla Hoff. 2006. \emph{{Poverty
Traps}}. Princeton University Press.

\leavevmode\hypertarget{ref-Bryman:2008va}{}%
Bryman, A. 2008. \emph{{Social Research Methods}}. 3 Ed. Oxford
University Press.

\leavevmode\hypertarget{ref-Castell:2020dg}{}%
Castell, Wolfgang zu, and Hannah Schrenk. 2020. {``{Computing the
adaptive cycle}.''} \emph{Scientific Reports} 10 (1): 1.

\leavevmode\hypertarget{ref-Clark:2020jd}{}%
Clark, William C, and Alicia G Harley. 2020. {``{Sustainability Science:
Toward a Synthesis}.''} \emph{Annual Review of Environment and
Resources} 45 (1): annurev-environ-012420-043621.

\leavevmode\hypertarget{ref-Anonymous:2016fv}{}%
Folke, Carl. 2016. {``{Resilience (Republished)}.''} \emph{Ecology and
Society} 21 (4): art44.

\leavevmode\hypertarget{ref-Griffiths:2004ey}{}%
Griffiths, Thomas L, and Mark Steyvers. 2004. {``{Finding scientific
topics.}''} \emph{P Natl Acad Sci Usa} 101 (suppl 1) (April): 5228--35.

\leavevmode\hypertarget{ref-Grun:2011tb}{}%
Grün, Bettina, and Kurt Hornik. 2011. {``{topicmodels: An R package for
fitting topic models}.''} \emph{J Stat Softw} 40 (13): 1--30.

\leavevmode\hypertarget{ref-Gunderson:2002vk}{}%
Gunderson, Lance H, and C S Holling. 2002. \emph{{Panarchy}}.
Understanding Transformations in Human and Natural Systems. Island
Press.

\leavevmode\hypertarget{ref-Holling:2001cu}{}%
Holling, C S. 2001. {``{Understanding the Complexity of Economic,
Ecological, and Social Systems}.''} \emph{Ecosystems} 4 (5): 390--405.

\leavevmode\hypertarget{ref-Holling:2002wu}{}%
Holling, C S, Lance Gunderson, and G Peterson. 2002. {``{Sustainability
and panarchies}.''} In \emph{Panarchy}. Island Press.

\leavevmode\hypertarget{ref-Holling:1996wx}{}%
Holling, C S, and G Meffe. 1996. {``{Command and control and the
pathology of natural resource management}''} 10 (2): 328--37.

\leavevmode\hypertarget{ref-biosphere:hJ4c_Ua8}{}%
Holling, CS. 1986. {``{The resilience of terrestrial ecosystems: local
surprise and global change}.''} In \emph{Sustainable Development of the
Biosphere}. Cambridge.

\leavevmode\hypertarget{ref-Parker:2012eo}{}%
Parker, John N, and Edward J Hackett. 2012. {``{Hot Spots and Hot
Moments in Scientific Collaborations and Social Movements}.''}
\emph{American Sociological Review} 77 (1): 21--44.

\leavevmode\hypertarget{ref-Redman:2003bd}{}%
Redman, Charles L., and Ann P Kinzig. 2003. {``{Resilience of Past
Landscapes: Resilience Theory, Society, and the Longue Dur{é}e}.''}
\emph{Conservation Ecology} 7 (1): art14.

\leavevmode\hypertarget{ref-Rocha:2015tk}{}%
Rocha, Juan Carlos, and Robin Wikström. 2015. {``{Detecting potential
impacts on ecosystem services related to ecological regime shifts {{}} a
matter of wording}.''} In \emph{Regime Shifts in the Anthropocene},
edited by Garry D Peterson and Reinette Oonsie Biggs. Stockholm:
Stockholm Resilience Centre, Stockholm University.

\leavevmode\hypertarget{ref-Zeng:2020hu}{}%
Zeng, Guanwen, Jianxi Gao, Louis Shekhtman, Shengmin Guo, Weifeng Lv,
Jianjun Wu, Hao Liu, et al. 2020. {``{Multiple metastable network states
in urban traffic}''} 117 (30): 17528--34.

\end{CSLReferences}

\pagebreak

\hypertarget{sec:SM}{%
\section{Supplementary Material}\label{sec:SM}}

\renewcommand\thefigure{S\arabic{figure}}
\renewcommand\thetable{S\arabic{table}}
\setcounter{table}{0}
\setcounter{figure}{0}

\begin{figure*}[h]
\centering
\includegraphics[width = 6in, height = 5in]{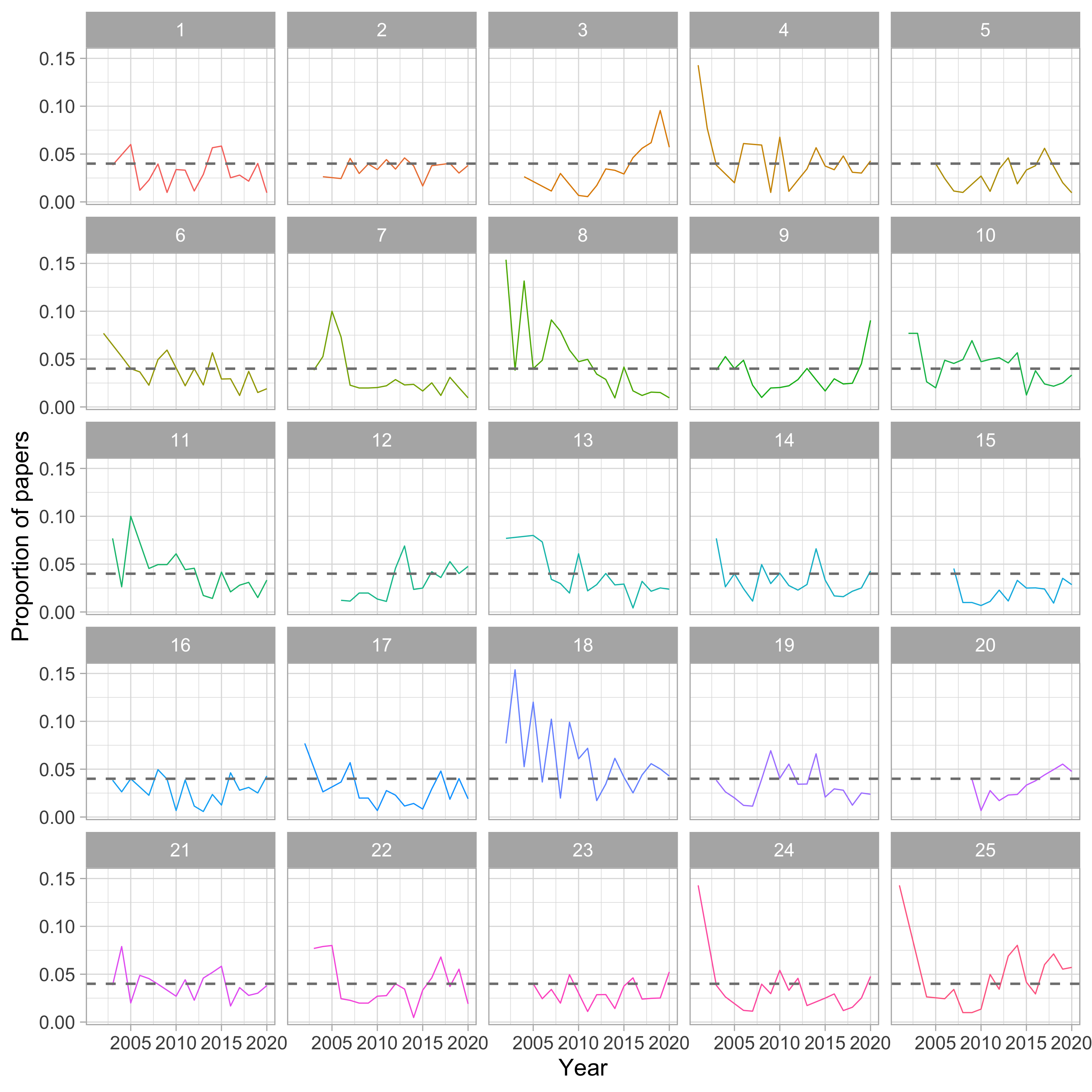}
\caption{\textbf{Proportion of papers per year per topic.} }
\label{fig:figS1}
\end{figure*}

\end{document}